# USING THE DEMPSTER SHAFER SCHEME IN A DIAGNOSTIC EXPERT SYSTEM SHELL

*Gautam Biswas and Tejwansh S. Anand*

Department of Computer Science
University of South Carolina
Columbia, S.C. 29208.

## ABSTRACT

This paper presents an application of the Dempster-Shafer evidence combination scheme in building a rule based expert system shell for diagnostic reasoning. Domain knowledge is stored as rules with associated belief functions. The reasoning component uses a combination of forward and backward inferencing mechanisms to interact with the user in a mixed initiative format.

## 1. INTRODUCTION

The development of expert systems like MYCIN and PROSPECTOR in the seventies led to the realization that their inferencing and the knowledge representation components could be separated from the domain-specific knowledge and applied to a completely different task. This led to the development of expert system shells like EMYCIN from MYCIN, KAS from PROSPECTOR, and EXPERT from CASNET. These shells or tools facilitate rapid development of the initial prototypes of an expert system. Each tool uses specific strategies for the representation of knowledge, inferencing, and overall control of the system. Therefore, each one addresses a specific class of problems. In addition to the ones mentioned above, a number of other developmental tools such as ROSIE, OPS5, ART, KEE and LOOPS are available now a days [1].

In this paper we present a diagnostic expert system shell that incorporates uncertain reasoning based on the Dempster Shafer framework. We describe the knowledge base structures employed and then elaborate on the inferencing mechanisms that integrate forward and backward chaining to enable the development of systems that incorporate mixed-initiative interactions with users. Section 2 briefly describes the Dempster Shafer evidence combination scheme. Section 3 discusses the knowledge base format and the control structures developed for the shell. Section 4 presents the conclusions of this paper.

## 2. INEXACT REASONING AND THE DEMPSTER SHAFER SCHEME

In many domains of application, the expert's reasoning processes are inherently uncertain, imprecise and incomplete. Moreover in complex real world situations the data gathering process is often incomplete and multiple sensors may provide mutually inconsistent data. It is this pervasive imprecision and uncertainty of the real world that requires the adoption of inexact reasoning processes in computer-aided decision making.

The traditional approaches to uncertain decision making based on Bayesian reasoning require a very large database of conditional probabilities, even if simplifying assumptions such as independence of individual observations are made though they may be hard to justify physically [2]. Shafer [3] demonstrated by simple examples, that the concept of ignorance is hard to represent in the Bayesian framework. The technique of assigning equal prior probabilities often produces counter intuitive results, especially when the number of hypotheses being considered is more than two. The MYCIN certainty factor scheme was proposed as an alternative to conditional probability schemes, but Adams [4] demonstrated that the combining functions can be derived from probability theory with the assumption of statistical independence. Most real life decision making involves complex problem solving in situations where facts and data available are insufficient, and knowledge of the domain is incomplete, therefore, a rigorous probabilistic analysis is not possible.

The drawbacks of the Bayesian and related adhoc schemes have drawn attention to the Dempster Shafer theory of evidence combination [5]. The key advantages of this theory are its abilities to explicitly incorporate the concept of ignorance in the decision making process, assign belief to subsets of hypotheses in addition to singleton hypothesis, and model the narrowing of the hypotheses set with the accumulation of evidence. This provides a better framework for modeling the human expert's reasoning process. Belief functions and their combining rules defined by the scheme are well suited to represent the incremental accumulation of evidence and the results of its aggregation.

The Dempster Shafer formulation is based on a frame of discernment, $\Theta$, a set of propositions or hypotheses about the exclusive and exhaustive possibilities in the domain under consideration. Two concepts that relate the impact of evidence on judgemental conclusions are the *measure of belief* committed exactly to a subset $A$ of $\Theta$ and the *total belief* committed to $A$. Exact belief relates to the situation where an observed evidence implies the subset of hypotheses, but this evidence does not provide any further discriminating evidence between individual hypotheses in $A$. The measure of **total** belief committed to a subset $A$ is defined as:

$$Bel(A) = \sum_{B \subseteq A} m(B), \tag{1}$$

where $m$ represents the exact belief function, and the summation is conducted over all $B$ that are subsets of $A$. Given a exact belief function $m$, if $\sum_{A \subset \Theta} m(A) < 1$, then $(1 - \sum_{A \subset \Theta} m(A))$ defines a measure of ignorance, denoted by $m(\Theta)$. In other words, $m(\Theta)$ is the extent to which the observations provide no discriminating evidence among the hypothesis in the frame of discernment $\Theta$.

Judgemental rules provided by experts basically represent individual pieces of evidence that imply subsets of hypothesis from $\Theta$ with belief values that correspond to exact belief functions (details of rule structure are given in section 3.1). Corresponding to two different pieces of evidence $e_1$ and $e_2$ with corresponding exact belief functions $m_1$ and $m_2$ over the same frame of discernment, Dempster's rule of orthogonal products is applied to combine the effects of observing the two pieces of evidence and compute a new exact belief function, $m$, that is given by

$$m(C_k) = \frac{\displaystyle\sum_{A_i \cap B_j = C_k} m_1(A_i) \, m_2(B_j)}{1 - \displaystyle\sum_{A_i \cap B_j = \varnothing} m_1(A_i) \, m_2(B_j)} \tag{2}$$

$A_i$ represents hypotheses subsets that are supported by $e_1$, $B_j$ represents hypotheses subsets supported by $e_2$, and $C_k$ represents the hypotheses subsets that are supported by the observation of both $e_1$ and $e_2$. The denominator is a normalizing factor to ensure that no belief is committed to the null hypothesis. More detailed discussions on the Dempster Shafer theory of evidence combination appear in [3,5].

## 3. THE EXPERT SYSTEM SHELL

Shells facilitate rapid prototyping and development of knowledge based systems. In addition they provide a convenient debugging tool for both knowledge engineers and domain experts. Most existing shells encode domain knowledge as rules, and inexact reasoning methods based on approximations of the Bayesian framework are integrated into the inferencing scheme. However, in spite of its advantages, to date no systems exist that adopt the Dempster Shafer scheme for evidence combination. We have developed OASES, a system for trouble shooting production processes [6] using this formalism. In the rest of this section we describe the current prototype of our expert system shell. The shell runs on an APOLLO DN 3000 under interpreted CommonLisp.

## 3.1. Knowledge Base Structure

The knowledge base of the diagnostic shell is in the form of a partitioned rule base. The structure of the rules (i.e., the rule language) as well as the partitions are designed in a manner that facilitate domain knowledge representation as well as efficient use of the Dempster Shafer evidence combination scheme.

The basic rule format is shown in Figure 1. A rule links a pattern on its left hand side (LHS) to one or more conclusions on its right hand side (RHS). The LHS pattern represents relevant evidence for the conclusions on the RHS. Single pieces of evidence are represented as attribute-value pairs, and, in general, a LHS pattern is a conjunction of pieces of evidence. Actually, evidence in the LHS pattern of a rule can be one of two types:

(i) **askable**, corresponding to evidence that can be obtained by directly querying the user (therefore, they have expert supplied queries associated with them), or

(ii) **verifiable**, corresponding to evidence obtained from rule firings. (These correspond to intermediate conclusions in the reasoning process).

Each conclusion on the RHS is a disjunctive set of hypotheses, where an individual hypothesis is represented as an attribute-value pair.

To accommodate uncertainty in the rule structure, there is an expert supplied belief function associated with the conclusions on the RHS. Belief functions are modeled as exact belief functions in the Dempster-Shafer framework. Note that belief values may also be associated with individual pieces of evidence in the LHS pattern. These may be derived from user input for askable patterns or computed values for verifiable patterns. *BF* is a Lisp function that computes the overall belief value for the LHS pattern of *a* rule. If multiple pieces of evidences are involved the belief values associated with the pattern is the *minimum* of the belief values of each piece of evidence on the LHS of the rule.

The expert may also supply evidence that negates the belief in a conclusion. An example from the OASES domain is:

**If** *(continuous flow fiberglass manufacture process) &*
*(molten glass viscosity is not nominal)&*
*(all ingredient compaction ratios are within limits)*
**then** *(rule out bin level fluctuations as the cause for the raw material sourcing problem).*

Such heuristic rules enable the expert to apply the process of elimination in the diagnostic process. In the D-S framework, evidence against a hypothesis is treated as evidence in favor of the negation of the hypothesis in the set theoretic sense. Therefore, if $\Theta$={*bin level fluctuations, inconsistency of raw materials, post scale contamination*} the above rule translates to

**If** *(continuous flow fiberglass manufacture process) &*
*(molten glass viscosity is not nominal)&*
*(all ingredient compaction ratios are within limits)*
**then** *(inconsistency of raw materials or post scale contamination*
*is the cause for the raw material sourcing problem).*

A simple rule editor has been developed that converts expert suplied rules into the internal Lisp format. It queries the expert for the LHS pattern of a rule and then the RHS conclusions. When the expert supplies rules which negate a hypothesis the appropriate conversion is done automatically. Each LHS pattern is treated as an independent piece of evidence.

As an example, consider the system designer entering an OASES rule. The rule editor prompts are depicted in bold font and the system designer's input is in italics.

**Enter the attributes and values of the LHS pattern:**
**(separate attributes and values by a comma; one attribute-value pair per line)**
*machinery speed and size, good balance*

**Enter the attributes and values of the RHS conclusions**
**(one conclusion per line; an attribute followed by a set of values separated by commas)**
*cause, materials management, workforce*
*cause, capacity planning, process design*

In order to extract the expert's belief in the RHS conclusions given the LHS evidence, the rule editor prompts the system designer to rank the RHS conclusions on a scale 1-10. Note here that the expert is not supplying an absolute support or belief value for the conclusion, but is merely providing a relative ranking based on his judgement.

**Enter the relative ranking for the conclusion on a scale 1-10.**
*9 3*

A very desirable feature of the D-S framework in representing knowledge in uncertain domains is the explicit definition and representation of ignorance. Therefore, in formulating the rule, the editor specifically queries the expert (system designer) as to his belief in the relevance of the LHS evidence, i.e., given that the system designer will be examining other evidence for making conclusions in this frame of discernment, on a scale of 1-10 to what extent does this contribute to making a final conclusion.

**On a scale of 1-10 what is the relevance of this evidence in the overall reasoning process?**
*8*

From this the system computes $m(\Theta)$ for the belief function corresponding to this rule to be 0.2 (i.e., 1 - 8/10). The relative ranking supplied by the expert is then normalized to yield the belief values (the $m$ function) for the rule according to the equation:

$$(bf)_{new} = (bf)_{old} \times \frac{1-m(\Theta)}{\sum(bf)_{old}}$$ (3)

Using $m(\Theta) = 0.2$, rule editor formulates the following rule:

[(<machinery_speed_and_size> <good_balance>) BF] →
        {[(<cause> <materials_management> <cause> <work_force>) 0.6]
        [(<cause> <capacity_planning> <cause> <process_design>) 0.2]}

The $m(\Theta)$ value along with the above values represents the measure of exact belief function for this rule.

    The rule editor also provides facilities for partitioning the knowledge base into separate chunks or units. This often facilitates modeling of the expert's reasoning process, and makes the knowledge acquisition process more structured, thus simplifying an otherwise difficult process. Conceptually, partitions may represent a successive refinement of the problem solving process. In complex domains, experts often follow a successive refinement process to keep the problem solving process manageable. For example, OASES [6], first identifies specific characteristics of the product and process, uses this information and general symptoms to establish a general cause category. Using the general cause as a frame of reference, it uses detailed information about the process and product, and observed deviations in performance to derive a specific cause. These partitions are sequential in nature in that the system descends from higher level partitions to lower level partitions successively.

    Another feature of the expert system shell is that it compiles rules into a **network**, to make the backward chaining phase of the inferencing more efficient, and to implement the chaining process in the Dempster-Shafer framework. This compilation is done off-line and the resultant network is stored as a multi-linked list. The rule network basically represents a compilation of

individual rules, and links conclusions to relevant evidence. Each sequential partition is compiled into a disjoint rule network. For example, Figure 2(b) shows a portion of the rule network corresponding to rules in XX [7], which deal with the identification of the site of a hydrocarbon play. The rules are listed in Figure 2(a). Both conclusions and evidence are represented in the same conceptual framework: attribute-value pairs. They form the nodes of the network. Rules relate evidence patterns to conclusions, and appear as links in the network. Links have weights associated with them. These weights are directly dependent on the amount of belief that the evidence pattern provides for the particular conclusion it is linked to. Final conclusions represent the top layer of nodes in the rule network. In Figure 2(b), the top layer of the network represent the plausible values of the site of a hydrocarbon play: *within the craton, on the continental shelf, and on the oceanic margin*. In addition there are two other kinds of *dummy* nodes. The first is an *AND* node: a conclusion that depends on a conjunction of evidences is linked to the evidence-nodes through this kind of node. The second is a *level* node. Level nodes link two hypotheses spaces. Conceptually, a hypothesis space is made up from the set of rules that verify the same attribute. It should be noted that the conclusions in a hypothesis space may be final conclusions or intermediate conclusions, and evidence may be either askable or verifiable. In Figure 2(b), the dashed box represents a hypothesis spaces. Nodes corresponding to verifiable evidence patterns are linked to level nodes and evidence nodes corresponding to the same attribute converge onto the same level-node. For example, in Figure 2(b) <beds_deepen> is a verifiable attribute, and therefore, all evidence nodes corresponding to the values of this attribute (*seaward* or *landward*) are linked to the same level-node. This level node is also linked to the hypothesis space that derives a value for this attribute. Thus, the overall structure of the rule network is that of hypotheses spaces linked to each other through level-nodes.

In inferencing terminology, verifiable pieces of evidence represent intermediate conclusions and their presence leads to chaining, or reasoning at multiple levels. To handle chaining in the Dempster Shafer framework each hypothesis space defines a separate frame of discernment (Θ). This approach closely mirrors the method suggested by Gordon and Shortliffe for implementing MYCIN [5] in this framework. The conceptual structures in the knowledge base are the attribute-value pairs and each verifiable attribute defines a frame of discernment extending over the possible values of that attribute. The main reason for such a implementation is to maintain mutual exclusiveness of hypotheses and independence of evidence in the Dempster Shafer framework.

In summary, the network links the hypotheses to the relevant evidence patterns based on the expert-supplied rules. Each RHS conclusion of a rule corresponds to a hypothesis-node, and is connected to evidence-nodes corresponding to the LHS pattern through weighted links. An evidence-node contains the attribute-value pair for that piece of evidence, and a pointer to the rule this evidence can fire. If the evidence is askable, the node contains a pointer to the associated query. For verifiable evidence, this pointer is null. Verifiable evidence-nodes are linked to level-nodes in the manner explained earlier [see Figure 2(b)]. The detailed description of the network and its method of implementation appears in [8]. The use of the network in inferencing is explained next.

### 3.2. Inferencing/Control Mechanism

The overall inferencing mechanism has four main components: the evidence combination scheme, the procedure for selecting the top ranked hypothesis, the query selection mechanism that directs the user-system dialogue based on the top ranked hypothesis, and a top level controller that is related to the selection of partitions within the rule base. The system adopts a mixed-initiative form of control. Based on evidence obtained, the system ranks its conclusions, and selects queries whose responses are likely to support the top ranked hypothesis. However, if the user considers the current query to be irrelevant, he/she may provide additional facts and other evidence, which causes the system to switch to the forward chaining mode. This illustrates mixed initiative control.

The overall flow of control for the inferencing mechanism is shown in Figure 3. The user interface is directed by the ASKQ routine. At startup, the system designer may require a specific set of questions be asked, or the user may be given the option to enter evidence relevant to the consultation. Based on the evidence obtained from the user's initial input, forward chaining and evidence combination using Dempster's combination formula (equation 2) is performed by the DEDUCE function. The function GETMAXH is invoked to select a leading hypothesis based on belief values for intermediate and final conclusions, and the backward chaining process is initiated. CHOOSEQ uses the top ranked hypothesis to select appropriate queries. User responses lead to selection and firing of rules by the DEDUCE function, and new belief values are computed. At each step, before querying the user for more information, the system checks if based on the current belief values it has come to a definite conclusion using an EXITCHK function defined by the system designer.

To illustrate the inferencing scheme, we present an example from the XX system [7]. The system is currently trying to establish the site of a hydrocarbon play. Initial evidence provided by the user results in the exact belief function:

$m(\{h_1, h_2\}) = 0.45$, $m(h_1) = 0.25$, $m(h_2) = m(h_3) = 0.1$

where, $h_1$ = (<site of play> <margin>), $h_2$ = (<site of play> <shelf>), and $h_3$ = (<site of play> <craton>). Based on these values, GETMAXH establishes $h_1$ as the leading hypothesis, and CHOOSEQ is invoked to pick an appropriate query to obtain more evidence from the user. CHOOSEQ examines the rule network (Figure 2(b)) and picks (<dist> <less_equal_200>) as the most appropriate evidence-node. Since this is an askable attribute, the system queries the user and determines the distance of the play from the margin is *less than 200 miles*. This causes rule03 (Figure 2(a)) to fire and belief values get updated as shown below.

| $m_{rule03}$ <br> $m$ | $\{h_1, h_2\}$ (0.8) | $\Theta$ (0.2) |
|---|---|---|
| $\{h_1, h_2\}$ (0.45) | $\{h_1, h_2\}$ (0.36) | $\{h_1, h_2\}$ (0.09) |
| $h_1$ (0.25) | $h_1$ (0.2) | $h_1$ (0.05) |
| $h_2$ (0.1) | $h_2$ (0.08) | $h_2$ (0.02) |
| $h_3$ (0.1) | $\Phi$ (0.08) | $h_3$ (0.02) |
| $\Theta$ (0.1) | $\{h_1, h_2\}$ (0.08) | $\Theta$ (0.08) |

The updated belief function is:

$m(\{h_1, h_2\}) = 0.576$, $m(h_1) = 0.272$, $m(h_2) = 0.109$, and $m(h_3) = 0.022$.

The GETMAXH function again identifies $h_1$ as the leading hypothesis. To further increase belief in $h_1$, CHOOSEQ determines it has to establish evidence corresponding to the *AND* node in Figure 2(b), which involves establishing two verifiable attributes. Let us assume that the system has established that there is *no abrupt change in slope*, and *as we move seaward* the *beds dip seaward*. It now descends to Level 2 to determine the *direction of deepening of the beds*. Again Figure 2(b) indicates a number of askable evidence patterns to establish that *beds deepen seaward*. CHOOSEQ first selects a query to determine the direction in which sediments become finer, and the user responds *seaward*. This establishes the hypothesis (<beds_deepen> <seaward>) with a belief value of 0.7. In a traditional aproach, this would trigger rule 06, and belief values of the <site_of_play> attribute would be recomputed. However, in this case, establishing additional properties, such as *homogeneity of sediments* and *deepening of fauna* increase belief in the hypothesis (<beds_deepen> <seaward>), and result in different belief value computations for the <site_of_play> attribute.

In order to avoid repeated computation of belief values at different levels which is expensive [9], the system does not propagate belief values from a hypothesis space till the EXITCHK conditions are satisfied. In this example, the system continues to query the user till the belief value for (<beds_deepen> <seaward>) becomes 0.96, and the EXITCHK conditions are satisfied. Only then is this value propagated to a higher level, where belief values are updated in the *site of play* frame of discernment.

## 4. CONCLUSIONS

In this paper we have demonstrated an application of the Dempster Shafer inexact reasoning scheme in an expert system shell. The inferencing mechanism combines forward and backward reasoning to provide mixed-initiative control. A simple rule editor facilitates the entry of domain-specific rules and the belief values associated with the conclusions. The rule network is used to select queries and rules during the backward chaining process and propagate belief values. A shortcoming of the current approach is that belief values are propagated from one level to another, only after the EXITCHK condition for the first level is satisfied. Therefore, the system does not completely utilize all the information it has to determine the top ranked hypothesis using which it then focuses its dialogue. To overcome this limitation, work is in progress to develop methods that will allow the propagation of incremental changes in belief values.

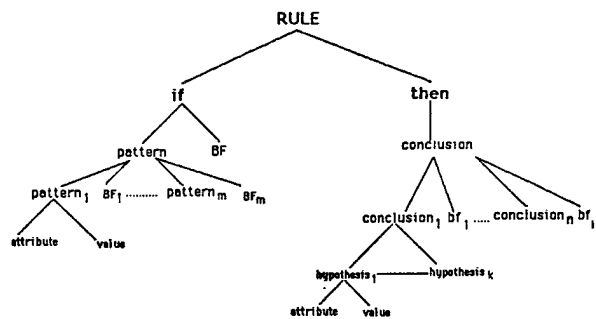

**Figure 1: The Rule Format**

```
(rule03
  (((<dist> <less_equal_200>)) BF)
  (((((<site of play> <shelf>)
      (<site of play> <margin>)) 0.8)) )

(rule04
  (((<dist> <greater_200>)) BF)
  (((((<site of play> <craton>)) 0.6)) )

(rule06
  (((<move> <seaward>)
    (<beds_dip> <seaward>)
    (<beds_deepen> <seaward>)
    (<abrupt_change> <no>)) BF)
  ((<site of play> <margin>)) 0.7)) )

(rule18
  (((<sed_finer> <seaward>)) BF)
  (((((<beds_deepen> <seaward>)) 0.7)) )

(rule19
  (((<sed_finer> <landward>)) BF)
  (((((<beds_deepen> <landward>)) 0.7)) )

(rule20
  (((<sed_homogenuous> <seaward>)) BF)
  (((((<beds_deepen> <seaward>)) 0.7)) )

(rule21
  (((<fauna_deepens> <seaward>)) BF)
  (((((<beds_deepen> <seaward>)) 0.7)) )

(rule21
  (((<reflectors_thin_&_dip> <seaward>)) BF)
  (((((<beds_dip> <seaward>)) 0.6)) )
```

**Figure 2(a): Some of the XX rules**

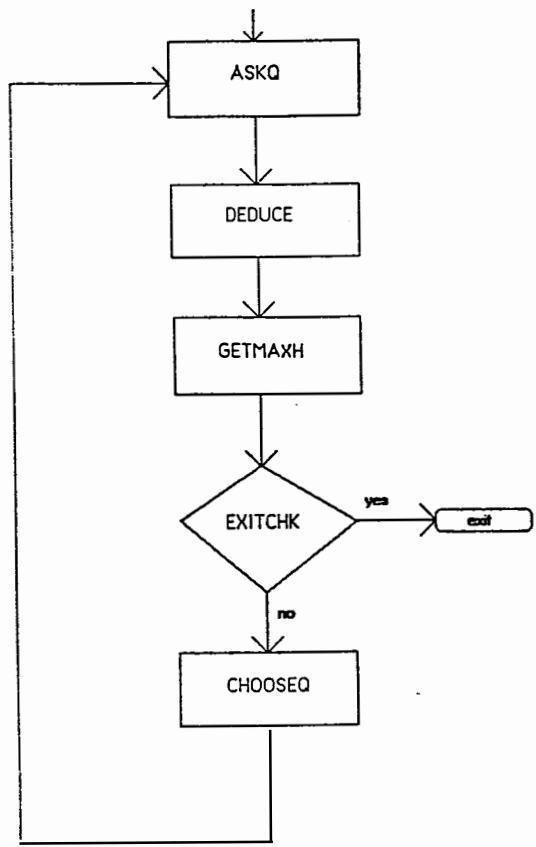

**Figure 3: The Inference Control Structure**

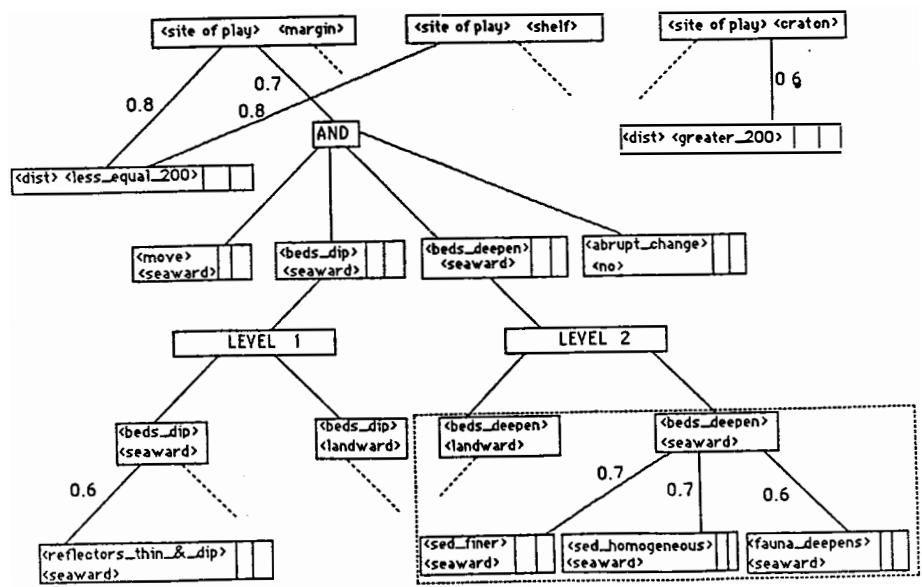

**Figure 2(b): A Section of the Rule Network for XX**

105